\pdfoutput=1

\documentclass[letterpaper, 10 pt, conference]{ieeeconf}  

\IEEEoverridecommandlockouts                              
\usepackage{amsmath,graphicx, algorithm, algorithmic,amssymb}
\usepackage{color}

\title{\LARGE \bf
MGCN: Semi-supervised Classification in Multi-layer Graphs with Graph Convolutional Networks 
}

\author{Mahsa Ghorbani, Mahdieh Soleymani Baghshah, Hamid R. Rabiee \\
 Computer Engineering Department, Sharif University of Technology \\ \{mahsa.ghorbani, soleymani, rabiee\}@sharif.edu 
}

\makeatletter
\newcommand\notsotiny{\@setfontsize\notsotiny\@vipt\@viipt}
\makeatother
\begin{document}

\maketitle
\thispagestyle{empty}
\pagestyle{empty}

\begin{abstract}

Graph embedding is an important approach for graph analysis tasks such as node classification and link prediction. The goal of graph embedding is to find a low dimensional representation of graph nodes that preserves the graph information. Recent methods like Graph Convolutional Network (GCN) try to consider node attributes (if available) besides node relations and learn node embeddings for unsupervised and semi-supervised tasks on graphs. 
On the other hand, multi-layer graph analysis has been received much attention more recently. However, the existing methods for multi-layer graph embedding can't incorporate all available information (like node attributes). Moreover, most of them consider either type of nodes or type of edges and don't treat within and between layer edges differently. In this paper, we propose a method called MGCN that utilizes the GCN for multi-layer graphs. MGCN embeds nodes of multi-layer graphs using both (within and between layers) relations and nodes attributes. 
We evaluate our method on the semi-supervised node classification task. Experimental results demonstrate the superiority of the proposed method to other multi-layer and single-layer competitors and also show the positive effect of using cross-layer edges.

\end{abstract}

\section{Introduction}
\label{sec:intro}
Graphs are powerful tools to model relations between entities like proteins, papers, people, and so on. Graph analytic gives us insight into hidden information in graphs that is useful for applications, such as link prediction \cite{wei2017cross}, node classification \cite{kipf2016semi}, and recommendation \cite{ying2018graph}. The most challenging problem in graph analysis is high computational and space cost and also the complexity of the graph space. Graph embedding provides an efficient way to solve these challenges by mapping the graph nodes into a low-dimensional space.

Although deep learning based methods have shown their capability in graph analysis, most of the existing methods only focus on single-layer graphs. However, in many real-world tasks, nodes and/or edges of the graphs have different types. Multi-layer graphs provide a framework to accommodate different types of entities and relations.  
In this paper, we intend to work on node embedding of multi-layer graphs containing some inter-connected single-layer graphs, where each of them contains nodes/edges of the same type. In these graphs, intra-layer edges (or within-layer edges) connect nodes with the same type and inter-layer edges (or between-layer edges) connect different types of nodes. 
According to the importance of multi-layer graphs in modeling real-world problems, we exploit recent advances in graph convolutional networks to embed nodes of these graphs more appropriately. We take advantage of the Graph Convolutional Network (GCN) in our framework to design a new node embedding method for multi-layer attributed graphs which is able to preserve input graph structure while also able to predict the label of nodes. The main contributions of the proposed method can be summarized as:
\begin{itemize}
\item 
It provides a framework for node embedding of multi-layer graphs using GCNs for the first time to the best of our knowledge. 
\item 
Due to the deep architecture, our method can be trained end-to-end as opposed to the most of the existing multi-layer graph embedding methods. 
\item 
The proposed loss function is composed of the structure reconstruction error and the classification error. It is also able to embed node attributes simultaneously.
\end{itemize}

\section{Related Works}
\label{sec:related}
Most of the graph embedding approaches that are based on deep learning methods can be divided into two categories. The first category includes methods that are based on the definition of filters and operators like convolution in the spectral and spatial domains of graphs \cite{defferrard2016convolutional,kipf2016semi}. In the second category, graph embeddings are learned by methods inspired from word embedding methods like Skip-Gram (they use sampled random-walks on the graph for this purpose) \cite{perozzi2014deepwalk,grover2016node2vec}. 

Almost all of the existing methods only work on single-layer graphs.
However, multi-layer graphs have received attention recently and different types of multi-layer graphs are introduced in \cite{kivela2014m}.
Most of the existing methods in this area have been designed for a particular type of multi-layer graphs and have restricting assumptions on the input, such as considering the type for either nodes or edges (and not for both of them) \cite{shanthamallu2018attention,xue2018deep}, considering the hierarchical structure or star-like structure between the layers \cite{zitnik2017predicting,ma2018multi,shi2018aspem}, and so on. These assumptions are restrictive for modeling some problems (e.g., the mutli-layer graph between papers, authors, and conferences). 

Recently, Li et al. \cite{li2018multi} claim that their method, called MANE, is the first method which considers between-layer edges in the optimization problem of node embedding (indeed, it considers types of nodes and edges in multi-layer graphs).
MANE has the most similar assumption about the input graph to our method, but it uses a different approach to solve the problem and is unable to incorporate attributes of nodes and also cannot learn node embedding and the classifier simultaneously via an end-to-end approach. 

\section{Graph Convolutional Network}
\label{sec:gcn}
Graph convolutional neural networks that have been introduced in the last few years provide powerful solution for combining node attributes and relations to obtain embedding for nodes of graphs. In this section, we focus on a specific type of spectral graph convolutional networks. 

Let $A\in \mathbb{R}^{N\times N}$ be the adjacency matrix of a graph with $N$ nodes and $D$ be a diagonal matrix containing node degrees.
$ L=I_N-D^{-\frac{1}{2}}AD^{-\frac{1}{2}}$ is called the normalized Laplacian matrix which is positive semi-definite and can be decomposed as $U\Lambda U^T$.
The main goal of GCN is to find a mapping from node attributes to a new space considering both the graph structure between the nodes and the available attributes for the nodes. For this purpose, columns of $U$ are chosen as the basis of the graph spectral space called the graph Fourier basis and $\Lambda$ contains Fourier frequencies. The goal is to design an appropriate filter in the graph spectral domain and convolve it with node attributes to map them into a new space. Finally, an inverse transform applies on the new attributes and maps them back to the graph spatial domain. 
Defferrard et al. \cite{defferrard2016convolutional} offer a truncated expansion in terms of Chebyshev polynomials to prevent eigen-decomposition of Laplacian matrix. Kipf et al. \cite{kipf2016semi} simplify the filter by limiting the order of Chebyshev polynomial and parameter sharing and propose a transformation on attributes of nodes (i.e. $x$) as $\theta\widetilde{D}^{-\frac{1}{2}}\widetilde{A}\widetilde{D}^{-\frac{1}{2}}x$, where $ \widetilde{A} = A + I_N$ and $\widetilde{D}_{ii}=\sum_j \widetilde{A}_{ij}$.

The generalized version of the above convolutional model with one hidden layer, $C$ attributes for each node ($X\in\mathbb{R}^{N\times C}$), and $F$ filters is defined as follows:
\begin{equation}
\label{calculateFeature}
    Z = a(\widetilde{D}^{-\frac{1}{2}}\widetilde{A}\widetilde{D}^{-\frac{1}{2}}X\Theta),
\end{equation}
where $\Theta\in \mathbb{R}^{C\times F}$ is the parameter matrix, $Z\in \mathbb{R}^{N\times F}$ is the new representation of nodes in the $F$-dimensional space \cite{kipf2016semi}, and $a$ is a non-linear activation function. This transformation has been called GCN layer that we utilize in our multi-layer graph embedding model. 

\section{Proposed Method for multi-layer Graph Embedding}
\label{sec:mgcn}
In this paper, we propose a framework for multi-layer graph node embedding by considering both within and between layer edges and also labels of a subset of nodes (to solve semi-supervised classification). For this purpose, we propose a deep architecture for mapping the nodes of the input multi-layer graph to a low-dimensional space via a loss function that is composed of two part. An unsupervised part that tries to preserve both within and between layer neighborhood structure of nodes in the embedding space. Meanwhile, when labels for a portion of nodes may be available, we can train more powerful representation for nodes. Therefore, we design also a supervised part to exploit available labels for a subset of nodes as the supervisory information. By jointly optimizing these loss functions, the learned embedding can preserve the structure and also be utilized to predict the class of nodes.

To design unsupervised part, we transform the features of nodes with the same type through a GCN layer according to within layer edges. So, for every layer of the input multi-layer graph, a GCN is utilized. It is worth noting that between-layer edges are not utilized in the GCN layers of our proposed method. However, we believe that representations of the connected nodes in different layers of a multi-layer graph are not independent due to between-layer connections. Therefore, we introduce a loss function that incorporates reconstruction of both the within-layer and between-layer connections from the obtained representation or embedding of nodes. It should be noted that the between-layer connections are employed just in the loss function, while the within-layer connections are used both in the loss function and in the GCN structure. On the other hand, to utilize the information about label of nodes, we add another GCN layer to predict the label of nodes from the node embeddings generated in the previous layer of the network. 

Let $G$ be the input multi-layer graph with $M$ layers where $G_k$ is the $k$-th layer with $N_k$ nodes. Also, let $A^{(k,l)}\in\mathbb{R}^{N_k\times N_l}$ be the adjacency matrix of the $k$-th layer, if $k=l$, and be the relation matrix between the $k$-th and the $l$-th layers, otherwise (the number of non-zero elements of $A^{(k,l)}$ is $|E^{(k,l)}|$).
On the other hand, assume that $Z^{(k)}\in\mathbb{R}^{N_k\times F_k}$ is the learned low dimensional representation of nodes for the $k$-th layer ($F_k$ is a pre-defined number that shows the embedding dimension and $F_k\ll N_k$). For simplicity, we assume that $F_k = F$ for all layers.

As we said before, for each layer of the multi-layer graph, we utilize a GCN to learn embedding of nodes that is capable of reconstructing both within and between layer edges. 
It has been shown that the inner product between the embedding of nodes is an appropriate decoder for reconstructing the relation between the nodes \cite{kipf2016variational}. 
Thus, we estimate the connection matrices as $\hat{A}^{(k,l)} = \sigma(Z^{(k)}Z^{(l)^T})$, where $k\ne l$ is used for the between-layer connections and $k=l$ for the within-layer connections ($\sigma(.)$ is the sigmoid function). 
To compare the reconstructed structure ($\hat{A}^{(k,l)}$) to the true structure ($A_{ij}$), 
the weighted binary cross-entropy is utilized:
\begin{equation}
\footnotesize
\begin{gathered}
    L_{link} =  -\sum_{k=1}^M \sum_{l=k}^M N_kN_l (\sum_{a^{(k,l)}_{ij}\in A^{(k,l)}} \frac{1}{|E^{(k,l)}|} a^{(k,l)}_{ij}\;log(\hat{a}^{(k,l)}_{ij})  + \\ \frac{1}{N_kN_l-|E^{(k,l)}|} (1-a^{(k,l)}_{ij})\;log(1-\hat{a}^{(k,l)}_{ij})),
\end{gathered}    
\end{equation}
where $L_{link}$ denotes the structure reconstruction capability of the obtained embeddings. 
Because of the graph sparsity, most elements of the target adjacency matrices are zeros and the resulting classification problem is imbalanced. To resolve it, we assign different balancing weights to the connected and disconnected pairs in the loss function.

\begin{figure}[!ht]
  \centering
  \centerline{\includegraphics[scale=0.205]{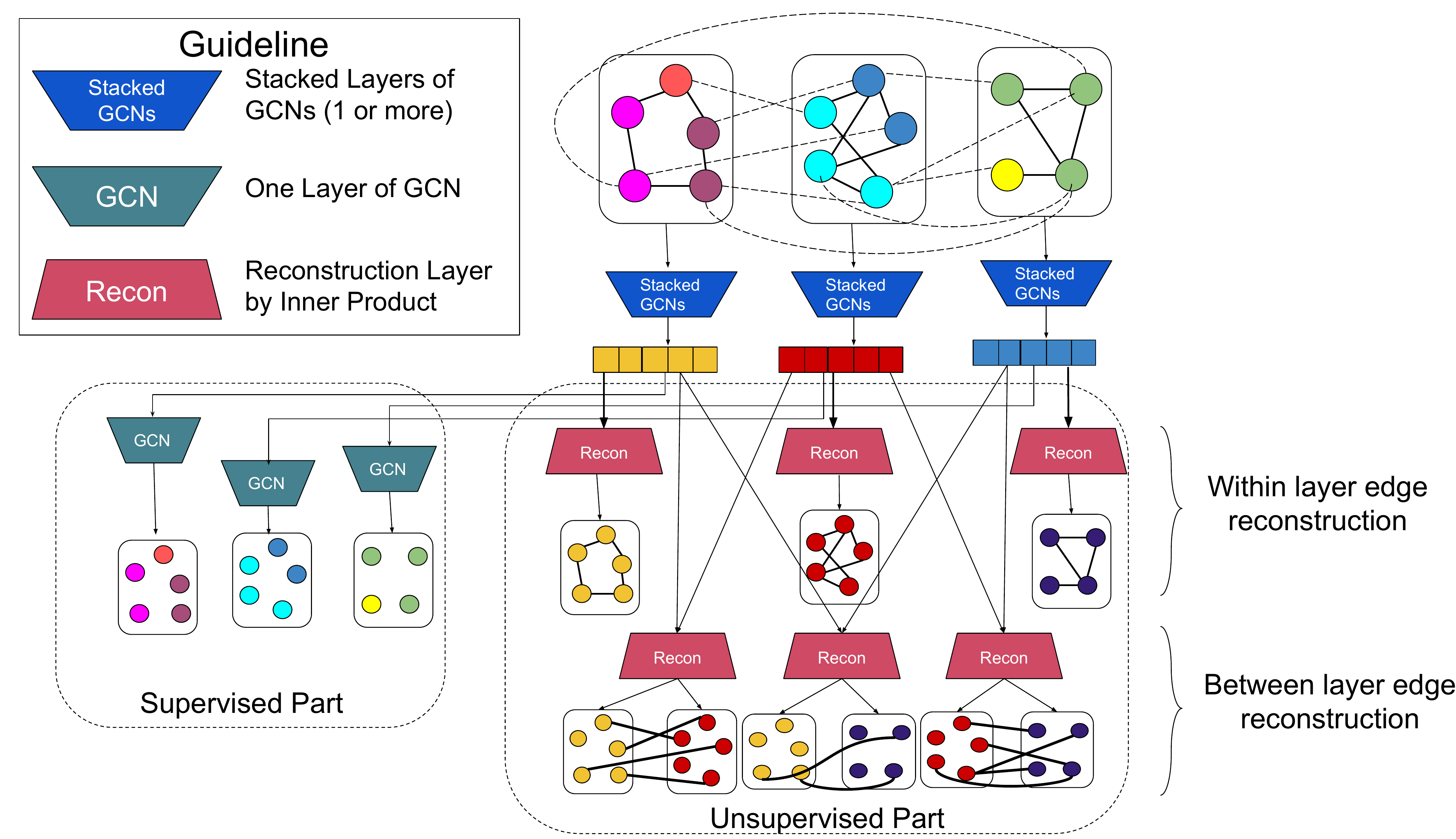}}
    \caption{MGCN contains an unsupervised part for within and between layer edge reconstruction and a supervised part for node label prediction.}
    \label{fig:mgcn}
\end{figure}

If we access the labels of at least some nodes, we can adapt the loss function to learn embeddings by a semi-supervised approach. In the semi-supervised classification, for each graph layer, another GCN is also employed on the obtained representations to predict labels (using a softmax activation) that enables end-to-end learning of representations for the semi-supervised classification.
The cross-entropy between the predicted labels and the true labels is used as the node classification loss function:
\begin{equation}
    L_{label} = - \sum_{l=1}^M \sum_{i \in S_l}\sum_{j=1}^{K_l} y^{(l)}_{ij} \; log(\hat{y}_{ij}^{(l)}),
\end{equation}
where $K_l$ is the number of classes for the nodes in the $l$-th layer, $S_l$ denotes the set containing indices of the labeled nodes in the $l$-th layer, and $y^{(l)}_{ij}$ and $\hat {y}^{(l)}_{ij}$ are the $j$-th element of vectors with the length $K_l$. In fact, $y^{(l)}_{i.}$ denotes a one-hot vector for the target label and $\hat{y}^{(l)}_{i.}$ shows the predicted probability of labels for the $i$-th node in the $l$-th layer of the graph.
We use a weighted sum of the above losses as:
\begin{equation}
\label{eq:loss}
    L_{total} = L_{link} + \lambda L_{label}.
\end{equation}
An overview of the MGCN is shown in Fig. \ref{fig:mgcn} and the steps of this algorithm are provided in Algorithm \ref{alg:MGCN}.

\begin{algorithm}[!ht]
\caption{MGCN Algorithm for multi-layer graphs.}
\label{alg:MGCN}
\footnotesize
\begin{algorithmic}[1]
\REQUIRE Adjacency matrices of (within/between) relation matrices ($A^{(k,l)}\in\mathbb{R}^{N_k\times N_l}$), attributes of nodes in all layers ($X_k$s), the coefficient of the reconstruction loss ($\lambda$)\\
\ENSURE Embeddings of nodes in all the layers (i.e. $\forall{k,} Z^{(k)}\in\mathbb{R}^{N_k\times F_k}$)
\STATE{Initialize all parameters $\Theta$ of the network}
\FOR{number of iterations} 
        	\STATE{For each graph layer $k$, find embedding of all of its nodes as \\ $Z^{(k)} \leftarrow$ GCN} 
            \STATE{For each pair of graph layers $k$ and $l$ ($k\leq l$), reconstruct (within/between) relation matrix as  $\hat{A}^{(k,l)}  \leftarrow \sigma(Z^{(k)}Z^{(l)^T})$}
            \STATE{For each graph layer $k$, find the label of its nodes as \\ $\hat{Y}^{(k)} \leftarrow$ softmax(GCN($Z^{(k)}$))}
	\STATE{Update all the parameters by gradient descent on $L_{total}$ (Eq.~\ref{eq:loss}): \\$\Theta \leftarrow \Theta - \nabla_{\Theta} (L_{total})$}
\ENDFOR
\end{algorithmic}
\end{algorithm}
\normalsize


\begin{table*}[!t]
\centering
\notsotiny
\caption{Results of models in semi-supervised node classification task on Infra and Aminer datasets.}
\label{tab:res}
\begin{tabular}{|c|c|c|c|c|c|c|c|c|c|c|c|c|}
\hline
Dataset                                                   & \multicolumn{6}{c|}{Infra}                                                 & \multicolumn{6}{c|}{Aminer}                                                                         \\ \hline
\begin{tabular}[c]{@{}c@{}}Training\\  Ratio\end{tabular} & \multicolumn{2}{c|}{20\%}       & \multicolumn{2}{c|}{50\%}       & \multicolumn{2}{c|}{80\%}       & \multicolumn{2}{c|}{20\%}       & \multicolumn{2}{c|}{50\%}       & \multicolumn{2}{c|}{80\%}       \\ \hline
Score                                                     & Micro-F1       & Macro-F1       & Micro-F1       & Macro-F1       & Micro-F1       & Macro-F1       & Micro-F1       & Macro-F1       & Micro-F1       & Macro-F1       & Micro-F1       & Macro-F1       \\ \hline
DeepWalk-Single                                           & 0.740          & 0.655          & 0.738          & 0.658          & 0.750          & 0.699          & 0.887          & 0.515          & 0.895          & 0.572          & 0.897          & 0.590          \\ \hline
DeepWalk-Hetero                                           & 0.582          & 0.436          & 0.613          & 0.475          & 0.629          & 0.504          & 0.870          & 0.383          & 0.894          & 0.431          & 0.899          & 0.447          \\ \hline
Node2vec-Single                                           & 0.854          & 0.719          & 0.854          & 0.727          & 0.866          & 0.757          & 0.898          & 0.522          & 0.902          & 0.546          & 0.905          & 0.577          \\ \hline
Node2vec-Hetero                                           & 0.612          & 0.481          & 0.643          & 0.507          & 0.655          & 0.506          & 0.901          & 0.437          & 0.918          & 0.482          & 0.921          & 0.495          \\ \hline
MANE                                                      & 0.874          & 0.599          & 0.881          & 0.607          & 0.880          & 0.606          & 0.904          & 0.500          & 0.921          & 0.530          & \textbf{0.926} & 0.534          \\ \hline
GCN                                                       & 0.736          & 0.479          & 0.794          & 0.583          & 0.885          & 0.618          & 0.916          & 0.769          & 0.911          & 0.777          & 0.893          & 0.733          \\ \hline
MGCN                                                      & \textbf{0.985} & \textbf{0.921} & \textbf{0.992} & \textbf{0.959} & \textbf{0.994} & \textbf{0.978} & \textbf{0.921} & \textbf{0.791} & \textbf{0.924} & \textbf{0.795} & 0.921          & \textbf{0.778} \\ \hline
\end{tabular}
\end{table*}
\normalsize
\section{Experiments}
In this section, we evaluate the proposed method on the semi-supervised node classification task.
\label{sec:exp}
\subsection{Datasets}
\label{ssec:dataset}
We use two datasets for evaluation of our method: Infrastructure system network (Infra) and academic collaboration network (Aminer). \\
\textbf{Infra} is a three layered graph which contains (1) an airport network, (2) an AS network, and (3) a power grid \cite{li2018multi}. All layers are functionally dependent and labels of nodes in the layers are based on the service areas inferred from the geographic proximity. Infra dataset contains $8,325$ nodes, $5,138$ within edges, $23,897$ between edges, and $5$ classes.\\
\textbf{Aminer} is also a three layered graph of academic collaboration in computer science which contains (1) paper (paper-paper citation), (2) people (co-authorship), and (3) conference venues (a venue-venue citation). There are two cross layers, who writes which paper and which venue publishes which paper. The node labels are based on the research areas. Aminer dataset contains $17,504$ nodes, $107,466$ within edges, $35,229$ between edges, and $5$ classes.

Since none of the compared methods is able to handle node attributes, the selected datasets don't have attribute on nodes and the identity matrix is used as node attributes for MGCN and GCN. 
\subsection{Experimental Setup}
\label{ssec:exp-setup}
To evaluate the performance of MGCN, the following methods are selected to be compared:\\
\textbf{MANE} \cite{li2018multi}: It has been designed for multi-layer graphs and is based on the eigen-decomposition of a matrix made up of within and between layer relation matrices.  \\
\textbf{DeepWalk} \cite{perozzi2014deepwalk}: Deepwalk is the baseline method for single-layer graphs and is based on the uniform random walk on the graph and usage of the Skip-Gram method. \\
\textbf{Node2vec} \cite{grover2016node2vec}: This method has also been designed for single-layer graphs and is the generalized version of DeepWalk which uses biased random walks.\\
\textbf{GCN} \cite{kipf2016semi}: This method uses GCN on each layer separately, without using cross edges where each GCN classifies nodes of one of the graph layers. 

We report the results for two versions of DeepWalk and Node2vec. In the first version (i.e. single), results of node classification in each layer are calculated and aggregated to compute the overall measure. In the second version (i.e. heterogeneous), the nodes and edges of graph layers are accumulated to built a new single layer graph. 

Since the compared methods (except to GCN) can't utilize labels during learning of node embeddings (i.e. they can obtain node embeddings only based on the graph structure), to evaluate them on semi-supervised node classification task, a logistic regression classifier is trained on the embeddings. In other words, after obtaining the embeddings without using label of nodes, these embeddings are considered as the input feature for the classifier and the labeled subset of nodes are used as training data to learn the classifier. Finally the learned classifier is used to predict labels of unlabeled subset of nodes. The parameters of the compared methods are set based on the suggestions in their papers. For Node2vec, $p$ is set to 1 and for $q$, four values $0.25, 0.5, 2\;\text{, and}\; 4$ are tested and the best results are reported. For MGCN, the parameter $\lambda$ in the loss function is selected by cross-validation to be 10 and ReLU is chosen as non-linear activation function. All the experiments are run 10 times and the average results are reported. The embedding dimension is set to 32.
\subsection{Results}
\label{ssec:results}
The Micro-F1 and Macro-F1 scores of the methods using different portions of training labels are reported in Table~\ref{tab:res}. 
By comparing the results for both versions of DeepWalk and Node2vec, we can conclude that dividing layers based on their types and analyzing them separately has better (or competitive) results than working with one accumulated graph. Comparing the results of MGCN and GCN shows that considering between-layer edges improves the results, especially the Macro-F1 values. On Aminer datasets, although the results of methods are so close in Micro-F1, the results of MGCN is better with a large margin in Macro-F1. 
Generally, the proposed MGCN outperforms the other methods significantly.

\begin{figure}[!ht]
  \centering
  \centerline{\includegraphics[width=9cm]{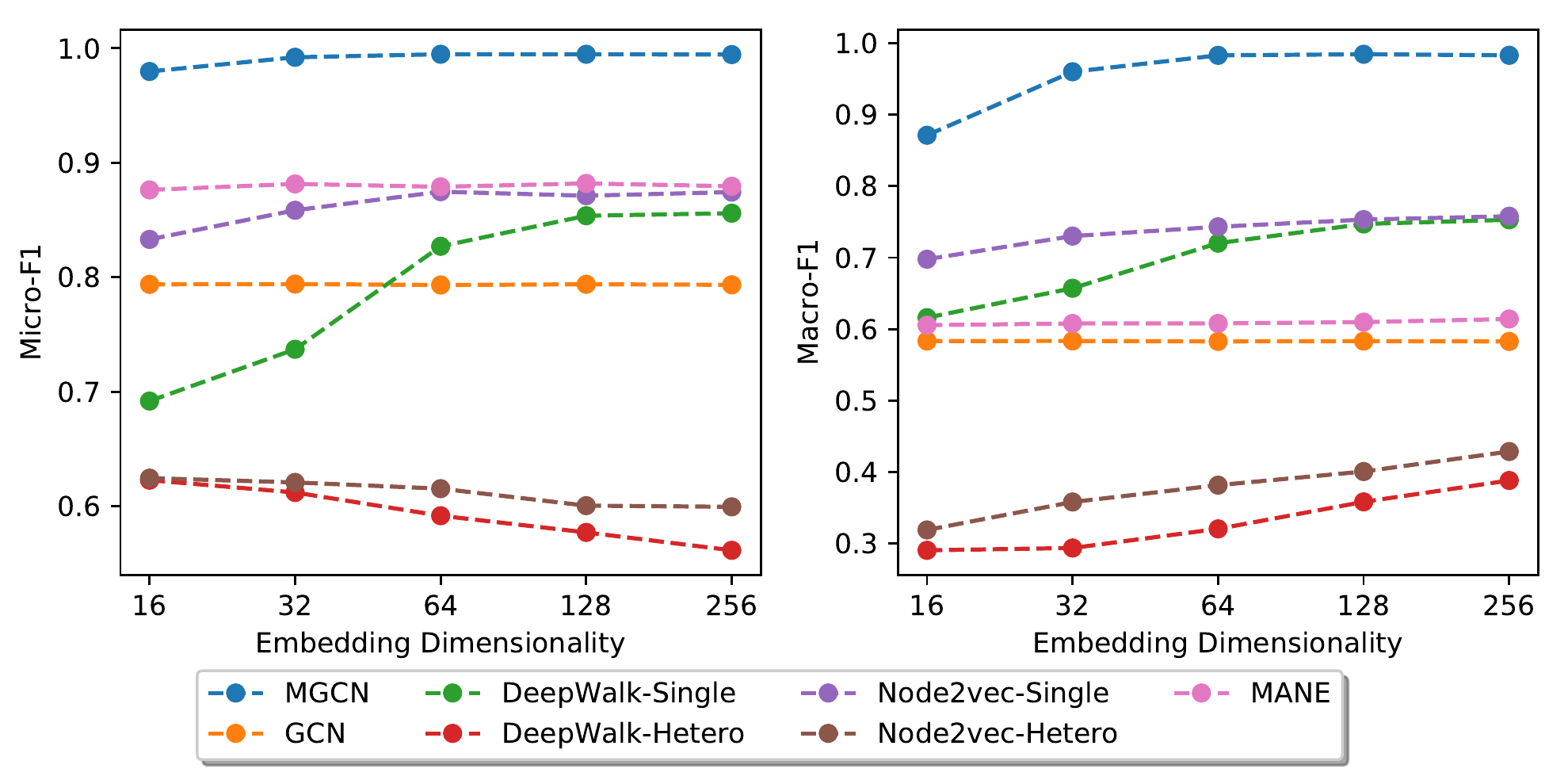}}
  \caption{The effect of embedding dimension on Infra dataset.}
  \label{fig:dimension}
\end{figure}

It's reasonable that increasing the embedding dimension makes the results better since the methods have more degree of freedom to embed nodes. According to Fig.~\ref{fig:dimension}, MGCN shows high performance in both measures even in low dimensional embedding spaces and its performance increases when the embedding dimension grows. 

\section{Conclusion}
GCNs are rapidly becoming more popular for non-Euclidean data embedding. However, embedding nodes of multi-layer graphs has not been studied so much, and especially graph convolution has never been applied on these graphs. In this paper, we extended the GCN model to embed multi-layer graph structure (and also node attributes when available) and propose an end-to-end deep learning method, named MGCN, that is able to find representation of nodes considering all available information for semi-supervised classification. We showed the superiority of MGCN in considering between-layer edges to some single-layer graph embedding methods and also to a recent multi-layer graph embedding method on the node classification task. 

\bibliographystyle{IEEEbib}
\bibliography{refs}

\end{document}